\documentclass[conference]{IEEEtran}
\IEEEoverridecommandlockouts
\usepackage{color}
\usepackage{graphicx}
\usepackage{xcolor}
\usepackage{amsmath}
\usepackage{soul}
\usepackage{url}
\usepackage{hyperref}

\usepackage[utf8]{inputenc}
\usepackage{pgfplots}
\pgfplotsset{compat=1.18}

\makeatletter
\def\hlinewd#1{%
\noalign{\ifnum0=`}\fi\hrule \@height #1 %
\futurelet\reserved@a\@xhline}
\makeatother
\usepackage{booktabs}
\usepackage{multirow}

\usepackage{siunitx}

\usepackage[algoruled,noline,longend,linesnumbered]{algorithm2e}

\usepackage{pgfplots}
\usepackage{pgfplotstable}
\usepackage{siunitx}
\makeatletter

\usepackage{amsthm}

\usepackage[small,compact]{titlesec} 
\titlespacing{\section}{2pt}{2pt}{2pt}
\titlespacing{\subsection}{2pt}{2pt}{2pt}
\titlespacing{\subsubsection}{2pt}{2pt}{2pt}

\begin{document}

\graphicspath{{Fig/}}
\def\figname{Figure}
\def\algname{Algorithm}

\newcommand{\papertitle}{Class-Aware Pruning for Efficient Neural Networks  
\vspace{-17pt} 
}

\title{\papertitle}

 \author{
 \IEEEauthorblockN{Mengnan Jiang$^1$,
 Jingcun Wang$^1$, Amro Eldebiky$^2$, Xunzhao Yin$^3$, Cheng Zhuo$^3$, Ing-Chao Lin$^4$, 
 Grace Li Zhang$^1$}
 \IEEEauthorblockA{$^1$TU Darmstadt, $^2$TU Munich, $^3$Zhejiang University $^4$ National Cheng Kung University}
 \IEEEauthorblockA{Email: \{mengnan.jiang, jingcun.wang, grace.zhang\}@tu-darmstadt.de, amro.eldebiky@tum.de, \\
xzyin1@zju.edu.cn, czhuo@zju.edu.cn, iclin@mail.ncku.edu.tw}
\vspace{-22pt} 
}

\maketitle

\begin{abstract}
\label{sec:abstract}
Deep neural networks (DNNs) have demonstrated remarkable success in various fields. However, the large number of floating-point operations (FLOPs) in DNNs poses challenges for their deployment in resource-constrained applications, e.g., edge devices. 
To address the problem, pruning has been introduced
to reduce the computational cost in executing DNNs. Previous pruning strategies are based on weight values, gradient values and activation outputs. Different from previous pruning solutions, in this paper, we propose a class-aware pruning technique to compress DNNs, which provides a novel perspective to reduce the computational cost of DNNs. 
In each iteration, the neural network training is modified to facilitate the class-aware pruning. Afterwards, 
the importance of filters with respect to the number of classes is evaluated. The filters that are only important for a few number of classes are removed. The neural network is then retrained to compensate for the incurred accuracy loss. 
The pruning iterations end until no filter can be removed anymore, indicating that the remaining filters are very important for many classes. 
This pruning technique outperforms previous pruning solutions in terms of accuracy, pruning ratio and the reduction of FLOPs. Experimental results confirm that this class-aware pruning technique can significantly reduce the number of weights and FLOPs,
while maintaining a high inference accuracy. Our code is available at \url{https://github.com/HWAI-TUDa/Class-Aware-Pruning} 
\end{abstract}

\section{Introduction} \label{sec:intro} 

Deep neural networks (DNNs) have achieved significant success in various fields, e.g., computer vision \cite{vit}, and natural language processing \cite{bert}. 
However, their significant success comes with an increasing number of parameters and multiply-accumulate (MAC) operations as well as floating-point operations (FLOPs). For instance, ResNet-50, a widely used convolutional neural network for image classification, contains 25.6 million parameters and requires approximately 4.1 billion MAC operations and thus 8.2 billion FLOPs to process an RGB image with 224$\times$224 pixels. 
The huge number of MAC operations and FLOPs prevents the application of DNNs in resource-constrained platforms, e.g., edge devices. 

To reduce the computational cost of DNNs, various techniques, e.g., pruning \cite{lee2020flexible}\cite{powerpruning},  
quantization \cite{spantidi2023automated}\cite{classquan}, and knowledge distillation \cite{jang2023pipe}\cite{de2021knowledge}, 
have been proposed at the software level. 
Pruning refers to the removal of unnecessary weights in DNNs according to a specific criterion. 
According to the granularity of pruning, it can be categorized into unstructured pruning and structured pruning. 
Unstructured pruning removes individual weights without considering their structures. For example, 
in \cite{unstructured0},  
weights with small absolute values are pruned.
\cite{unstructured7} proposes a weight pruning technique that removes weights based on the product of weight values and their gradients.
\cite{unstructured4} adjusts the neural network training to punish the number of non-zero parameters, so that more weights can be reduced to zero after training. 
\cite{grasp} iteratively prunes the weights that have low products of the their values, gradients, and the second-order gradients.
 
 Unstructured pruning can achieve a high pruning rate. However, the weight matrix after unstructured pruning tends to be irregular, which is not efficient for digital hardware. 
To address this challenge, structured pruning has been introduced to prune groups of weights, such as filters \cite{Depgraph}, layers \cite{layerpruning}, 
or weight blocks, e.g., residual blocks in ResNet \cite{wu2018blockdrop}\cite{Resnet}. 
Among structured pruning, filter-wise pruning is a widely used pruning technique since it provides a relatively fine granularity for compressing DNNs.
To prune filters, criteria such as weights \cite{lin2022pruning}\cite{TPP}, 
activation outputs \cite{HRank}\cite{chip}, gradients \cite{TaylorScore}, and specific parameters \cite{SCP} are usually used to guide the pruning process. For example, \cite{L1} remove the weight of filters based on their absolute values. \cite{trimming} calculates the average percentage of zero activation outputs of filters and removes those filters that generate activation outputs with a large percentage of zeros. \cite{gradient1} prunes filters based on both activation outputs and their gradients. 

In this paper, we propose a class-aware pruning technique to compress DNNs. 
In this technique, we evaluate the importance of each filter with respect to the number of classes. 
By removing filters that are important for only a few classes, the number of FLOPs in DNNs can be reduced significantly while still maintaining high inference accuracy. 
The main contributions of this work are summarized as follows:
\begin{itemize}
     \item This paper proposes a class perspective to prune filters. Instead of using the values of weights, gradients, or activation outputs, the importance of filters with respect to the number of classes is used as a novel
     criterion to compress DNNs.
     \item To facilitate the class-aware pruning, we propose to modify neural network training to push the network to generate a clear differentiation between important and unimportant filters. 
     \item The importance of filters with respect to the number of classes is quantitatively defined. With this importance indicator, filters that are important for only a few classes are pruned iteratively. The network is then retrained to compensate for the incurred accuracy loss.



    \item Experimental results demonstrate that the proposed class-aware pruning method can reduce the number of FLOPs by up to 77.1\% while still maintaining a high inference accuracy. In addition, 
    compared with other pruning methods, the proposed method achieves a higher accuracy with a larger pruning rate and the reduction of FLOPs.
    
\end{itemize}

The rest of this paper is organized as follows. Section \ref{sec:PreliminariesMotivation}
describes the background and motivation of this work. Section \ref{sec:methods} explains the evaluation of the importance of filters and the pruning strategy. Section \ref{sec:results} and \ref{sec:conclusion} present the experimental results and conclusions, respectively. 





\section{Background and Motivation}
\label{sec:PreliminariesMotivation}

\subsection{Background}
\label{sec:Preliminaries} 
Pruning 
removes unnecessary weights and thus reduce the number of MAC operations and FLOPs. 
Unstructured pruning removes weights without considering their structures. The resulting sparse weight matrix is not friendly for hardware platforms, e.g., systolic array in Google's TPU, to execute, since a lot of zero weight values still need to be processed on hardware or additional hardware overhead is required to skip such zero values \cite{zhang2021hardware}. Different from unstructured pruning, structured pruning considers the structures of weight matrices for pruning. 
For example, filter-wise pruning removes whole filters. Accordingly, such a pruning strategy is friendly for hardware implementations.

Previous work on structured pruning such as filter-wise pruning removes weights of filters typically based on three criteria, namely weight values, the values of the activation outputs generated by the filters and gradients of weights. For example, 
\cite{L1}\cite{TPP} remove the filters based on the absolute values of their weights.
\cite{Depgraph} prunes filters based on the sum of square roots of their weights. 
\cite{lin2022pruning} eliminates filters based on both the absolute weight values and the reduction of FLOPs incurred by this filter pruning.
\cite{SSS} introduces filter-wise regularization and adjusts the training to generate more filters with zero values.

Different from weight-based criterion, 
activation output criteria are motivated by the output feature maps generated by filters. 
For example,  \cite{trimming} evaluates the average percentage of zeros in activation outputs and removes those filters that generate activation outputs with a large percentage of zeros.
\cite{HRank} determines the rank of feature maps using the singular value decomposition (SVD) and removes those filters that generate low-rank feature maps. Similar to \cite{HRank}, \cite{chip} evaluates the 
values of the singular values of feature maps in the SVD decomposition and removes filters that generate feature maps with low singular values.

Gradient-based criterion evaluates filters of weights through gradients derived from the back-propagation process in training. 
For example, \cite{gradient1}  prunes filters based on both activation outputs and their gradients.
 Unlike \cite{gradient1}, \cite{gradient2} prunes filters based on the average product of weights and their gradients within a filter.  \cite{gradient3} uses scaling factors and their gradients for filter pruning while minimizing scaling factors in training.



\subsection{Motivation}
 \label{sec:Motivation}

\begin{figure}
    \centering
    \includegraphics[]{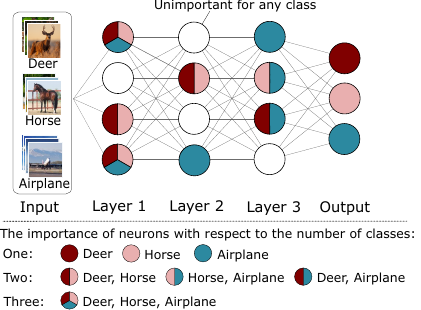}
    \caption{ An example of the importance of neurons for different number of  classes. 
    }
    \label{fig:pathway}
\end{figure}

Different from previous pruning solutions, 
we introduce a novel class-aware perspective to prune filters. \textit{The class is defined as a set of input images sharing the same label, e.g., cat.} The basic concept of this class-aware pruning is that neurons/filters contribute to different numbers of classes in image classification since images of different classes trigger different neuron paths across layers \cite{int2018}. For example, a neuron/filter can be important for the classification of several classes since it locates on the neuron paths of these classes.
Another neuron/filter can be important for only one class, since it is located only on the neuron/filter path of this class.

 Figure \ref{fig:pathway} illustrates the class-aware pruning concept, where 
 a fully-connected neural network with four layers is used to classify three classes, deer, horse and airplane. 
Assume that the neuron importance with respect to the number of classes has been evaluated.
It can be seen from this figure 
some neurons contribute only to one of the three classes, while other neurons are important for two classes and even three classes. The neurons with white color contribute to none of the classes and can be pruned completely. It might also be possible to prune those neurons that are only important for one class and adjust the neural network by retraining, so that the remaining neurons can  compensate for the accuracy loss of this pruning. 
With this class-aware pruning, the remaining neurons in the network are important for at least two classes, indicating that they are useful for image classifications. In this figure, neurons are used as an example to explain their importance differences with respect to the number of classes. However, the class-aware concept can also be applied to filter-wise pruning. 



\section{ Proposed Class-Aware Pruning Method}
\label{sec:methods}
In this section, the proposed class-aware pruning method is presented. 
The pruning process begins with a modified training of neural networks for facilitating class-aware pruning, as described in Section III-A. 
The importance scores of filters with respect to the number of classes are then evaluated in Section III-B. 
After this evaluation, filters with low importance scores will be removed iteratively. Neural networks are then fine-tuned to compensate for the incurred accuracy loss in each pruning iteration. The pruning and fine-tuning strategies are described in Section III-C. 
 The overall class-aware pruning framework is provided in Section III-D.


\begin{figure}
    \centering
    \includegraphics[]{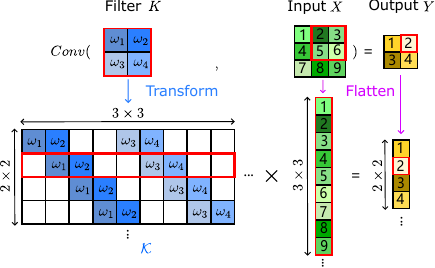}
    \caption{ Reshaping weights of filters in
a convolutional layer to match the flattened input feature map.
    }
    \label{fig:OrthConv}
\end{figure}
\subsection{Modified Neural Network Training}
\label{sec: NN_training}
To train a neural network to facilitate the class-aware pruning, we modify the cost function as follows. This modified cost includes three parts, namely the original cross entropy $\mathcal{L_{CE}}$, $\mathcal{L}_{1}$ regularization and the orthogonality term $\mathcal{L}_{orth}$:
\begin{align}
\label{eq:loss}
& \mathcal{L} = \mathcal{L_{CE}} + \lambda_{1}\mathcal{L}_{1} + \lambda_{2}\mathcal{L}_{orth}\\
& \mathcal{L}_{1} = \sum_{l=1}^{H} {\left\| {\mathbf{W}_l} \right\|_{1}}, \quad \mathcal{L}_{orth} = \sum_{l=1}^{H} {\left\| \mathcal{K}\mathcal{K}^T - I \right\|_{2}}
\end{align}
where 
$\lambda_{1}$ and $\lambda_{2}$ are coefficients for $\mathcal{L}_{1}$ and 
$\mathcal{L}_{orth}$, respectively. ${\mathbf{W}_l}$ is the weight matrix in the $l$th layer. $H$ and $I$ are the number of layers in the network and an identity matrix, respectively. $\mathcal{K}$ is a transformed matrix for weights of filters in a convolutional layer.
An example of weight matrix transformation is illustrated in 
Figure~\ref{fig:OrthConv}, where the convolution operations between a filter $K$ with a size $1\times 2 \times 2$
and an input $X$ with a size $ 3 \times 3$ are performed with a stride of one. 
To use each input value in $X$ once in this convolution process, 
the filter $K$ 
is transformed to a $4 \times 9$ sparse matrix $\mathcal{K}$, 
where each row represents the original filter sliding over different positions on the input data. With a stride of one, the offset for consecutive rows is one. 
Multiplying $\mathcal{K}$ with the flattened input generates the same output as the convolution operation.



The motivation for using $\mathcal{L}_{1}$ regularization is to push neural networks to generate sparse weight matrices. The filters with many zero values are usually not important for many classes, so that they can be pruned. Different from $\mathcal{L}_{1}$ regularization, $\mathcal{L}_{orth}$ aims to push neural networks to learn orthogonal filters that can capture diverse features, which has been deployed in \cite{OrthConv} to enhance the inference accuracy of neural networks. 
Such orthogonal filters can be useful for many classes. By combining these two regularization terms, the training of neural networks will be adjusted to generate a polarized importance score distribution of filters, facilitating a clear differentiation between important and unimportant filters.


\subsection{ Evaluation of Importance Scores for Filters}

To quantitatively evaluate how many classes a filter is important for, we first estimate the importance score of this filter with respect to one class, e.g., cat. The importance scores of the filter with respect to all classes are added together as the 
total importance score.

To evaluate the importance score of a filter with respect to a specific class, a given number of images of this class in the training data are randomly selected. Such images are then used to derive the sensitivities of activations in the output feature map generated by this filter to the cost function of the pre-trained neural network. 
If their sensitivities are low, 
the importance of this filter with respect to this class can be considered as low. 

To derive the sensitivity of an activation output generated by the filter, 
we set this activation value to 0 and verify the change of the cost function of the pre-trained neural network for images of this class. For example, given the $f$th filter and an input image $x_j$ which belongs to the $n$th class, e.g., cat,    
in training dataset, the importance of the $i$th activation output $a_i^f$ 
can be evaluated as follows: 
\begin{equation}
\label{eq:importance_path}
\Theta(a_i^f,x_j)
=\left|\mathcal{L}(x_j)-\mathcal{L}\left(x_j ; 
a_i^f \leftarrow 0\right)\right|
\end{equation}
where 
${a}_i^f \leftarrow 0$ indicates that the activation ${a}_i^f$ is set to zero. 
$\Theta(a_i^f,x_j)$ represents the importance of this output activation.  

\begin{figure}
    \centering
    \includegraphics[]{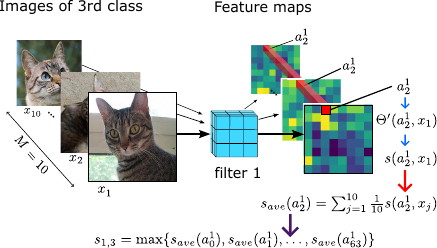}
    \caption{
    An example of evaluating the importance score for one activation output $a_{2}^{1}$ in the 1st filter for the $3$rd class, namely cat, where $2$ represents the second activation output.
    Images ($x_1^{3}, x_2^{3}, \ldots, x_{10}^{3}$) in the $3$rd class are the inputs for the network. 
    }
    \label{fig:howtoscore}
\end{figure}

The importance of the remaining activation outputs generated by this filter can be evaluated similarly. However, this evaluation 
requires a forward inference for each individual activation output, which is time-consuming in practice. 
To improve the computational efficiency, 
the importance score evaluation 
is approximated  
using a first-order Taylor expansion as follows. This method evaluates the importance scores of all activation outputs generated by the filter by 
performing a single backward propagation only once: 
\begin{equation}
\label{eq:taylor}
\Theta^\prime(a_i^f,x_j)
=\left|a_i^f \frac{\partial \mathcal{L}(x_j)}{\partial a_i^{f}} \right|
\end{equation}
where $\Theta^\prime(a_i^f,x_j)$ is defined as the Taylor-score of an activation output. If the Taylor-score of an activation output is near zero, this activation can be considered 
not to
contribute significantly to this class. 
Accordingly, we define the \textit{importance score} of the activation output to the $n$th class as follows: 

\begin{equation}
\label{eq:con_taylor}
s(a_i^f,x_j) = \begin{cases}
1,\  \Theta^\prime(a_i^f,x_j)> \tau \\
0,\  \Theta^\prime(a_i^f,x_j)\leq \tau
\end{cases}
\end{equation}
where $\tau$ is a threshold and was set to $10^{-50}$ in the experiment.

To make the importance score more general, we select several images of this class in the training dataset and derive the average importance score of each activation output as follows:
\begin{equation}
\label{eq:score}
s_{ave}(a_i^f)=
\sum_{j=1}^{M} \frac{1}{M} s(a_i^f,x_j) 
\end{equation}
where $M$ is the selected number of images of this class in the training dataset.


Equation~\ref{eq:score}  evaluates the importance score of only one activation output generated by the filter. The importance scores of the remaining activation outputs generated by the $f$th filter can be evaluated similarly. \textit{The importance score of the $f$th filter to the $n$th class is defined as 
the maximum importance score of the activation outputs as follows: }
\begin{equation}
\label{eq:thr_taylor}
s_{f,n} = \max \{ s_{ave}(a_0^f), s_{ave}(a_1^f), \ldots, s_{ave}(a_Z^f) \}
\end{equation}
where $Z$ is the number of activation output values generated by the filter.

The derivation of the importance score of a filter to a class is illustrated in Figure~\ref{fig:howtoscore}, where $M$ images of the cat class are selected to evaluate the importance score of a filter. The output activation $a_2^1$ is used as an example to illustrate the evaluation of the importance score. The importance score of the filter is the maximum value of the scores of activation outputs generated by the filter. 

\begin{figure}
    \centering
    \includegraphics[]{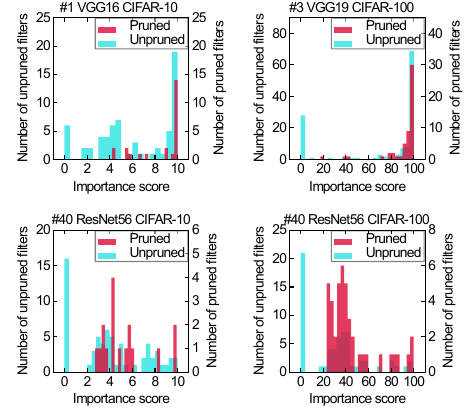}
    \caption{Distribution of filter importance scores before and after pruning in single layer. VGG16-CIFAR10: displayed is the first convolutional layer. VGG19-CIFAR100: displayed is the third convolutional layer. ResNet56-CIFAR10/100: displayed is the 40th convolutional layer.
    }
    \label{fig:shift}
\end{figure}

The importance score of a filter with respect to the remaining class can be evaluated in the same way. 
The total score of a filter 
with respect to all classes is the sum of the importance score with respect to each individual class. Figure~\ref{fig:shift} illustrates the distribution of importance scores of filters in one layer of VGG16, VGG19 and ResNet56 before pruning. According to this figure, it is clear that a large number of filters are not important for many classes so that they can be pruned to compress neural networks.





\subsection{Pruning and Fine-Tuning Strategies}



After evaluating the importance scores of filters with the method described above, those filters with low importance scores should be pruned. 
Specifically, we remove those filters with importance scores below a given threshold. Since different datasets vary in the number of classes, different thresholds should be used, e.g., 3 for CIFAR10 with 10 classes and 30 for CIFAR100 with 100 classes. 
To maintain a fine pruning granularity, 
we also restrict the pruning percentage in each iteration, e.g., no more than 10\%.

After removing those filters with low importance scores in one iteration, the neural network is fine-tuned with the modified cost function in Equation~\ref{eq:loss}. Afterwards, the importance scores of filters are reevaluated and pruned iteratively. The pruning iterations end until no filters can be pruned anymore. Figure~\ref{fig:shift} illustrates the distribution of importance scores of filters after pruning. Compared with the distribution before pruning, we can see that those filters unimportant for many classes are pruned. In addition, the remaining filters have higher scores, indicating that they are important for many classes.


\begin{figure}
    \centering
    \includegraphics{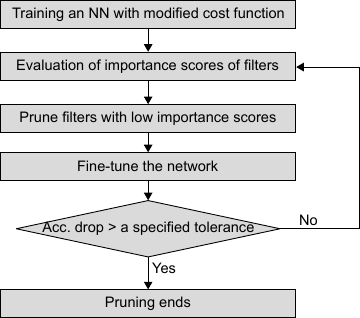}
    \caption{The proposed class-aware pruning framework.}
    \label{fig:framework}
\end{figure}
\subsection{The Overall Pruning Framework}

The overall class-aware pruning framework is summarized and illustrated in Figure~\ref{fig:framework}, where a neural network is pruned iteratively to remove those filters that are only important for a few classes. 
This framework starts with training a neural network with a modified cost function for facilitating class-aware pruning. To prune unimportant filters in this neural network, we first evaluate the importance score of each filter with respect to the number of classes. Afterwards, a filter pruning strategy is applied to efficiently remove those unimportant filters. 

The filter pruning described above might degrade the inference accuracy of neural networks. To recover their inference accuracy, the neural networks are fine-tuned by retraining. 
After the fine-tuning, if the accuracy degradation is smaller than a given threshold, another pruning iteration starts where 
the neural networks will be further pruned by reevaluating the importance scores of filters. 
In case that the accuracy cannot be recovered after the fine-tuning, the pruning iterations terminate.

\section{Experimental Results}\label{sec:results}
To verify the effectiveness of the proposed method, three neural networks and the corresponding datasets were used, namely, VGG16, 
 VGG19, 
and ResNet56
on CIFAR10/100. 
To evaluate the importance scores of filters for different classes, e.g., cat or dog, 10 images for each class were randomly selected in the  CIFAR-10/100 training datasets. We have verified that by evaluating more than 10 images the importance scores of filters are almost the same with those with 10 images. 
The optimizer used in training neural networks was Stochastic Gradient Descent (SGD), where a learning rate of 0.01 was initialized. The batch size was set as 256, weight decay as 0.0005, and momentum as 0.9. 
$\lambda_1$ was set to 0.0001 and $\lambda_2$ was set to 0.01 in Equation 1. 
During the iterative pruning process, retraining was performed for up to 130 epochs after each pruning iteration.
All experiments were conducted on an NVIDIA A100 80GB GPU.



\begin{table}
    \footnotesize
    \centering
    \caption{Pruning results with the proposed pruning method}
    \vspace{-2pt}
    \begin{tabular}{ccccc}
        \hline
         \multirow{2}{*}{NN-Dataset} &  \multicolumn{2}{c}{Acc. comp.}&  \multicolumn{2}{c}{ Pruning perf.}\\
         \cmidrule(lr){2-3} \cmidrule(lr){4-5}
          & Original & Pruned & Prun. ratio & FLOPs red. \\
         \hline
         VGG16-CIFAR10  & 93.90\% & 92.99\% & 95.6\% & 77.1\% \\
         VGG19-CIFAR100 & 73.49\% & 72.56\% & 85.4\% & 75.2\% \\
         ResNet56-CIFAR10 & 93.71\% & 92.89\% & 77.9\% & 62.3\% \\
         ResNet56-CIFAR100 & 72.36\% & 71.49\% & 50.0\% & 43.8\% \\
         \hline
    \end{tabular}
    \label{tab:drop1}
\end{table}


\begin{figure}
    \centering
    \includegraphics[]{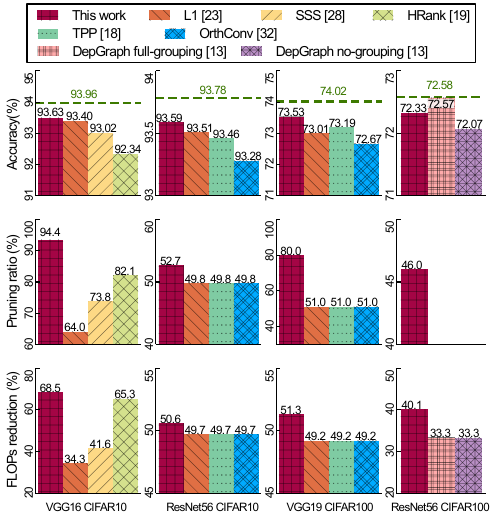}
    \caption{Comparisons of Top-1 accuracy, pruning ratio and reduction of FLOPs. The green dash line is the original accuracy of the neural network. The previous methods included in the comparison are L1 \cite{L1}, SSS \cite{SSS}, HRank \cite{HRank}, TPP \cite{TPP}, OrthConv \cite{OrthConv}, and DepGraph full-grouping \cite{Depgraph} and DepGraph no-grouping \cite{Depgraph}. }
    \label{fig:comp}
\end{figure}

Table~\ref{tab:drop1} demonstrates the performance of the proposed pruning technique. The first column shows the neural networks and datasets. The original Top-1 accuracy of such neural networks and their accuracy with the proposed pruning technique are compared in the second and the third columns. The original accuracy of neural networks are obtained by training neural networks from scratch. 
According to such two columns, there is only a slight inference accuracy loss. The last two columns show the pruning ratio and the reduction of the number of FLOPs with the proposed pruning technique. These two columns confirm that the proposed pruning technique can significantly 
reduce the computational cost. For example, for VGG16-CIFAR10, 95.6\% of the parameters are removed, so that the number of FLOPS is reduced by 77.1\%. For ResNet56, to ensure the shortcut connections during pruning, only the first layer of each residual block is pruned. Even under this constraint, 77.9\% of the parameters in ResNet56 are removed while a reduction of 62.3\% in FLOPs is achieved.

To demonstrate the advantages of the proposed pruning technique, 
we compared the accuracy  after pruning, the pruning ratio and the reduction of the number of FLOPs between the proposed method and the state-of-the-art solutions including L1 \cite{L1}, SSS \cite{SSS}, TPP \cite{TPP}, HRank \cite{HRank}, OrthConv \cite{OrthConv}, and DepGraph \cite{Depgraph}. To fairly compare the accuracy with different pruning techniques, 
we used the
pre-trained model weights 
from the previous work and applied the proposed pruning framework on neural networks. Accordingly, the original accuracy and the accuracy obtained with the proposed framework are a little bit different from that in Table I. The comparison results are shown in 
Figure~\ref{fig:comp}, where different neural networks are compared with those from different techniques due to the limited available results in the previous work. 
The top figure compares the accuracy, according to which the proposed pruning technique achieves the highest accuracy in most cases. The figures in the middle and at the bottom compare the pruning ratio and the reduction of the number of FLOPs, demonstrating the advantages of the proposed  method. 

\begin{figure}
    \centering
    \includegraphics[]{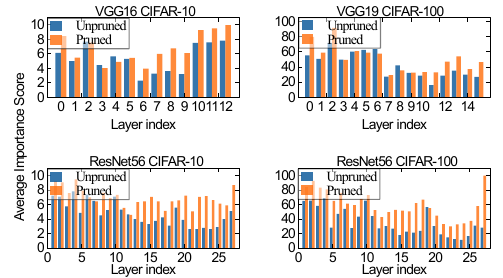}
    \caption{Average importance scores of filters before and after pruning in all layers of neural networks. 
    }
    \label{fig:mean}
\end{figure}

Since the proposed pruning technique removes those filters with low importance scores in an iterative way, the remaining filters after this pruning should have higher important scores compared with those in the original neural network, indicating they are very critical for many classes. To verify this, we compared the average importance scores of filters of neural networks before and after the proposed pruning technique. The results are shown in Figure~\ref{fig:mean}. 
It is obvious that for most layers, there is a considerable growth in importance scores after pruning. This shows that filters that are important only for a small number of classes are pruned, while remaining filters are important for many classes. 

\begin{table}
    \footnotesize
    \centering
    \caption{Performance of ResNet56 CIFAR10 under different  pruning strategies}
    \vspace{-5pt}
    \begin{tabular}{ccccc}
        \hline
        \multirow{2}{*}{Pruning strategy} &  \multicolumn{2}{c}{ Acc.}&  \multicolumn{2}{c}{ Pruning perf.}\\
         \cmidrule(lr){2-3} \cmidrule(lr){4-5}
          & Pruned & Drop & Prun. ratio & FLOPs red.\\
         \hline
        percentage & 92.76\% & -0.95\% & 73.7\% & 55.2\% \\
        threshold & 92.78\% & -0.94\% & 72.2\% & 60.4\% \\
        percentage+threshold & 92.89\% & -0.82\% & 77.9\% & 62.3\% \\
        \hline
    \end{tabular}
    \label{tab:strategy}
\end{table}
The proposed pruning technique used both importance score threshold and the pruning percentage as the pruning strategy. To demonstrate the advantage of this combination, we separately used each one as the individual pruning strategy and ran the pruning framework. The resulting accuracy drop and the pruning performance are demonstrated in Table~\ref{tab:strategy}. 
According to this table, 
it is clear that the combination of the importance score threshold and the pruning percentage has the best performance.

\begin{figure}
    \centering
    \includegraphics[]{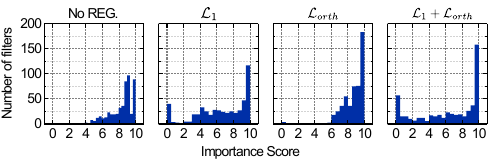}
    \caption{Importance score distribution of filters using different regularization strategies for VGG16 on CIFAR-10. }
    \label{fig:penalty}
\end{figure}

\begin{table}
    \footnotesize
    \centering
    \caption{Performance comparison with different cost functions}
    \vspace{-5pt}
    \begin{tabular}{ccccc}
        \hline
        \multirow{2}{*}{REG.} &  \multicolumn{2}{c}{ Acc.}&  \multicolumn{2}{c}{  Pruning perf.}\\
         \cmidrule(lr){2-3} \cmidrule(lr){4-5}
          & Pruned & Drop & Prun. ratio & FLOPs red. \\
         \hline    
         \multicolumn{5}{c}{\textit{VGG16-CIFAR10 original Acc. 93.90\%}} \\
        / & 92.91\% & -0.99\% & 73.6\% & 58.7\% \\         
        $\mathcal{L}_{1}$ & 93.06\% & -0.84\% & 91.8\% & 71.3\% \\
        $\mathcal{L}_{orth}$ & 93.10\% & -0.80\% & 74.5\% & 64.7\%  \\
        $\mathcal{L}_{1}$+$\mathcal{L}_{orth}$ & 93.16\% & -0.74\% & 94.8\% & 71.8\% \\
         \hline              
          \multicolumn{5}{c}{\textit{ResNet56-CIFAR10 original Acc. 93.71\%}} \\
        / & 92.74\% & -0.97\% & 69.4\% & 55.8\% \\         
        $\mathcal{L}_{1}$ & 92.77\% & -0.94\% & 72.0\% & 57.6\% \\
        $\mathcal{L}_{orth}$ & 92.73\% & -0.98\% & 69.3\% & 54.1\% \\
        $\mathcal{L}_{1}$+$\mathcal{L}_{orth}$ & 92.89\% & -0.82\% & 77.9\% & 62.3\% \\           
        \hline
    \end{tabular}
    \label{tab:penalty}
\end{table}

During training neural networks, we used both $\mathcal{L}_{1}$ and $\mathcal{L}_{orth}$ as the regularization strategies. 
To demonstrate the advantage of such a combination, we compared the importance score distributions of filters between 
 no regularization, $\mathcal{L}_{1}$, $\mathcal{L}_{orth}$, as well the combination of $\mathcal{L}_{1}$ and $\mathcal{L}_{orth}$.
The results are shown in 
 Figure~\ref{fig:penalty}.
 According to this figure, 
 $\mathcal{L}_{1}$ regularization led to a larger number of filters with an importance score of 0, while $\mathcal{L}_{orth}$ produced more high-score filters, especially those with a score of 10. Using both regularization strategies results in a more polarized importance score distribution of filters, facilitating a clearer distinction between important and unimportant filters. 
 Table~\ref{tab:penalty}, displaying the pruning results of neural networks using different regularization strategies, further confirms that the combination of $\mathcal{L}_{1}$ and $\mathcal{L}_{orth}$ achieves better performance compared with using individual $\mathcal{L}_{1}$ as well as $\mathcal{L}_{orth}$.


\section{Conclusion}
\label{sec:conclusion}
In this paper, we proposed a class-aware pruning technique to reduce the computational cost of executing neural networks.
By iteratively removing filters that are only important for a few classes, 
the neural networks become more compact while only 
keeping
filters 
very critical for many classes in the neural networks. Experimental results on several neural networks demonstrated that the pruning ratio of neural networks can be boosted up to 95.6\% and the number of FLOPs can be reduced by up to 77.1\%. 

\section*{Acknowledgement}
The work is supported in part by the Deutsche Forschungsgemeinschaft (DFG, German Research Foundation) -- Project-ID 457473137, in part by the National Science and Technology Council, Taiwan (R.O.C.) under grants NSTC 109-2628-E-006-012-MY3, 110-2221-E-006-084-MY3, in part by design services from Taiwan Semiconductor Research Institute (TSRI), Taiwan (R.O.C.), in part by the NSFC (Grant No. 62034007, 62141404) and in part by Major Program of Zhejiang Provincial NSF (Grant No. D24F040002) and SGC Cooperation Project (Grant No. M-0612).


\let\oldbibliography\thebibliography
\renewcommand{\thebibliography}[1]{%
\oldbibliography{#1}%
\fontsize{6.3pt}{6.3}\selectfont
\setlength{\itemsep}{0.06pt}%
}

\bibliographystyle{IEEEtran}
\bibliography{IEEEabrv,CONFabrv,bibfile}

\end{document}